\documentclass[preprint,12pt]{elsarticle}

\usepackage{graphicx}
\usepackage{subfig}

\usepackage{color}

\usepackage{amsmath}

\usepackage{lineno}

\journal{}

\begin{document}

\begin{frontmatter}


\title{IoU-balanced Loss Functions for Single-stage Object Detection}


\author[mymainaddress]{Shengkai Wu}
\ead{ShengkaiWu@hust.edu.cn}
\author[mymainaddress]{Jinrong Yang}
\ead{yangjinrong@hust.edu.cn}
\author[mysecondaryaddress]{Xinggang Wang}
\ead{xgwang@hust.edu.cn}
\author[mymainaddress]{Xiaoping Li\corref{mycorrespondingauthor}}
\cortext[mycorrespondingauthor]{Corresponding author}
\ead{lixiaoping@hust.edu.cn}
\address[mymainaddress]{State Key Laboratory of Digital Manufacturing Equipment and Technology, Huazhong University of Science and Technology, Wuhan, 430074, China. }
\address[mysecondaryaddress]{School of EIC, Huazhong University of Science and Technology, Wuhan, 430074, China.}

\begin{abstract}
Single-stage object detectors have been widely applied in computer vision applications due to their high efficiency. However, we find that the loss functions adopted by single-stage object detectors hurt the localization accuracy seriously. Firstly, the standard cross-entropy loss for classification is independent of the localization task and drives all the positive examples to learn as high classification scores as possible regardless of localization accuracy during training. As a result, there will be many detections that have high classification scores but low IoU or detections that have low classification scores but high IoU. Secondly, for the standard smooth L1 loss, the gradient is dominated by the outliers that have poor localization accuracy during training. The above two problems will decrease the localization accuracy of single-stage detectors. In this work, IoU-balanced loss functions that consist of IoU-balanced classification loss and IoU-balanced localization loss are proposed to solve the above problems. The IoU-balanced classification loss pays more attention to positive examples with high IoU and can enhance the correlation between classification and localization tasks. The IoU-balanced localization loss decreases the gradient of examples with low IoU and increases the gradient of examples with high IoU, which can improve the localization accuracy of models. Extensive experiments on challenging public datasets such as MS COCO, PASCAL VOC and Cityscapes demonstrate that both IoU-balanced losses can bring substantial improvement for the popular single-stage detectors, especially for the localization accuracy. On COCO \textit{test-dev}, the proposed methods can substantially improve AP by $1.0\%\sim1.7\%$ and $\text{AP}_{75}$ by $1.0\%\sim2.4\%$. On PASCAL VOC, it can also substantially improve AP by $1.3\%\sim1.5\%$ and $\text{A}{{\text{P}}_{80}}$, $\text{A}{{\text{P}}_{90}}$ by $1.6\%\sim3.9\%$. The source code will be made publicly available.
\end{abstract}


\end{frontmatter}


\section{Introduction}
\label{S:1}
Along with the advances in deep convolutional networks, lots of object detection models have been developed. All these models can be classified into single-stage detectors \cite{liu2016ssd,redmon2016you,lin2017focal,zhang2018single_RefineDet,zhang2018single_Entiched,li2019gradient} and multi-stage detectors \cite{ren2015faster,cai2018cascade,he2017mask,lin2017FPN,dai2016R-FCN,girshick2015fast,girshick2014r-cnn}. Improving the localization accuracy of object detection models is a challenging topic and many methods such as Cascade R-CNN \cite{cai2018cascade}, RefineDet \cite{zhang2018single_RefineDet} have been proposed to realize this goal by attaching more complex subnetworks which will hurt the efficiency of models. In this work, we aim to improve the localization accuracy of models without sacrificing efficiency. We find that the classification and localization loss functions adopted by most of the detection models are not good enough for accurate localization and the localization ability can be substantially improved by designing better loss functions that make no changes to the model's architecture. There are two problems with the loss functions adopted by most of the object detectors.

Firstly, the correlation between classification and localization task is weak. Most of the object detectors adopt the standard cross-entropy loss for classification which is independent of the localization task and this kind of classification loss will drive the model to learn as high classification scores as possible for all the positive examples regardless of their localization accuracy during training. As a result, the predicted classification scores will be independent of the localization accuracy and there will be many detections that have high classification scores but low IoU or detections with low classification scores but high IoU. These detections having the mismatch problem between the classification score and localization accuracy will hurt the performance of models in the subsequent procedure during inference. One the one hand, when traditional non-maximum suppression (NMS) or it's variants such as Soft-NMS \cite{bodla2017softnms} is applied, there will be cases that the detections with high classification scores but low IoU suppress the ones with low classification scores but high IoU. On the other hand, during computing COCO AP, all the detections are ranked based on the classification scores and there will be cases that the detections with high classification scores but low IoU are ranked ahead of the detections with low classification scores but high IoU, which will decrease the average precision. As a result, we claim that enhancing the correlation between classification and localization task is important for accurate localization. Secondly, during training, the gradients of localization loss for object detectors are dominated by outliers, which are examples with poorly localization accuracy. These outliers will prevent the models from obtaining high localization accuracy during training. Fast R-CNN \cite{girshick2015fast} proposes smooth L1 loss to suppress the gradients of outliers to a bounded value and can prevent exploding gradients effectively during training. However, the domination of outliers' gradients still exists during training and it is important to make more suppression on the gradient of outliers while increasing the gradient of inliers.

\begin{figure}[h]
\centering
\includegraphics[height=0.5\linewidth]{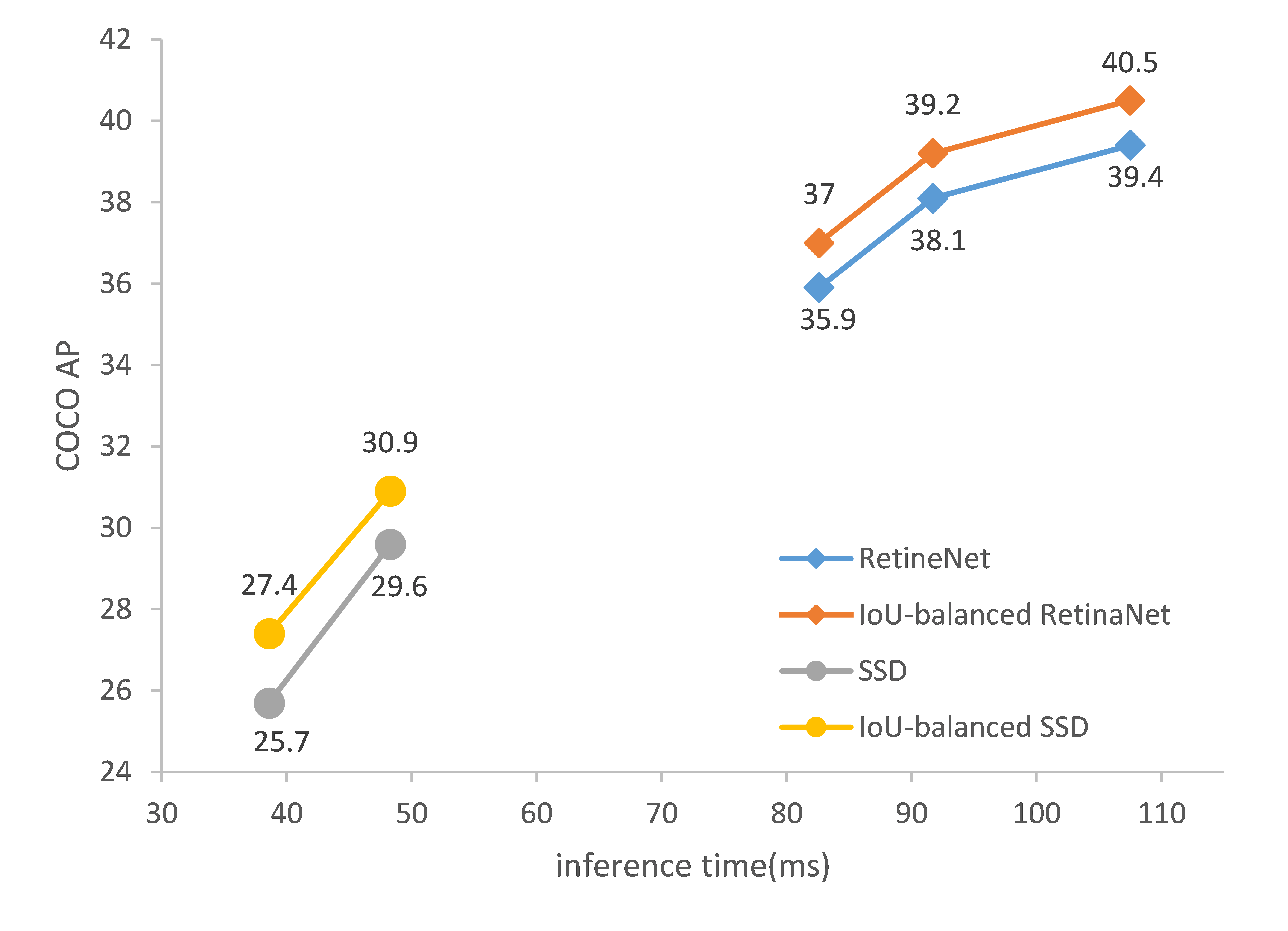}
\caption{Speed (ms) versus accuracy (AP) on COCO test-dev. IoU-balanced losses can consistently improve performance of different popular single-stage object detectors without sacrificing the inference time.}
\label{fig:speed versus accuracy}
\end{figure}

In this work, we propose IoU-balanced loss functions which consist of IoU-balanced classification loss and IoU-balanced localization loss to improve the localization accuracy of models. IoU-balanced classification loss pays more attention to positive examples with high IoU. The higher the IoU of the positive example is, the more contribution to the classification loss it makes. Thus, the positive examples with higher IoU will generate higher gradients during training and are more likely to learn higher classification scores. On the contrary, the positive examples with lower IoU are more likely to learn lower classification scores. This method will enhance the correlation between the classification and localization task. IoU-balanced localization loss up-weights the gradients of examples with high IoU while suppressing the gradients of examples with low IoU, making the model more powerful for accurate localization. Sufficient experiments on the challenging datasets such as MS COCO, Pascal VOC and Cityscapes demonstrate that IoU-balanced loss functions can substantially improve the performance of single-stage detectors without sacrificing efficiency as Figure \ref{fig:speed versus accuracy} shows. In addition, IoU-balanced losses can also improve the performance of multi-stage detectors, but the improvement is not as large as that for the single-stage detectors. It's because that for multi-stage detectors, the proposals generated by the first stage detector such as RPN are more accurate than the human-designed anchors in the single-stage detectors and the problem mentioned above is alleviated.

Our main contributions are as follows: (1) We demonstrate that the standard cross-entropy loss for classification and the smooth L1 loss for localization can hurt the localization accuracy of models and the localization ability can be substantially improved by designing better loss functions. (2) We propose IoU-balanced classification loss to enhance the correlation between the classification and localization tasks, which can substantially improve the performance of single-stage detectors. (3) We introduce IoU-balanced localization loss to up-weight the gradients of inliers while suppressing the gradients of outliers, which makes the models more powerful for accurate object localization.

The rest of this paper is organized as follows. Section \ref{S:2} introduces the related research work. Section \ref{S:3} introduces the IoU-balanced classification loss and IoU-balanced localization loss in details. Section \ref{S:4} presents sufficient experiments on several challenging datasets and demonstrates the effectiveness of our methods. Conclusions are given in Section \ref{S:5}

\section{Related Work}
\label{S:2}
\textbf{Accurate object localization}. Accurate object localization is a challenging topic for object detection and many methods to improve localization accuracy have been proposed in recent years. Multi-region detector \cite{gidaris2015multiregion} argues that a single regression step is insufficient for accurate localization and thus proposes iterative bounding box regression to refine the coordinates of detections, followed by NMS and box voting. Cascade R-CNN \cite{cai2018cascade} trains multi-stage R-CNNs with increasing IoU thresholds stage-by-stage and thus the multi-stage R-CNNs are sequentially more powerful for accurate localization. RefineDet \cite{zhang2018single_RefineDet} improves one-stage detector using two-step cascade regression. The ARM first refines the human-designed anchors and then the ODM accepts these refined anchors as inputs for the second stage regression, which is beneficial for improving localization accuracy. All these methods add new modules to the detection models and thus hurt efficiency. On the contrary, IoU-balanced loss functions improve localization accuracy without changing models' architecture and don't affect the efficiency of models.

\textbf{Hard example mining}. To improve the models' ability of handling hard examples, many hard example mining strategies having been developed for object detection. RPN \cite{ren2015faster} defines the anchors whose IoU with ground truth boxes are not larger than 0.3 as hard negative examples. Fast R-CNN \cite{girshick2015fast} defines the proposals that have a maximum IoU with ground truth boxes in the interval [0.1, 0.5) as hard negative examples. OHEM \cite{shrivastava2016OHEM} computes losses for all the examples, then ranks examples based on losses, followed by NMS. Finally, the top-B/N examples are selected as hard examples to train the model. SSD \cite{liu2016ssd} defines anchors whose IoU is lower than 0.5 as negative examples and ranks negative examples based on losses. The top-ranked negative examples are selected as hard negative examples. RetinaNet \cite{lin2017focal} designs focal loss to solve the extreme imbalance between easy examples and hard examples, which reduces the losses of easy examples whose predicted classification score is low and focuses more attention on hard examples whose predicted classification score is high. Libra R-CNN \cite{pang2019libraRCNN} constructs a histogram based on IoU for negative examples and selects examples from each bin in the histogram uniformly as hard negative examples. Different from these strategies, IoU-balanced loss functions don't change the sampling process and only assign different weights to the positive examples based on their IoU.

\textbf{Correlation between classification and localization task}. Most of the detection models adopt the parallel classification and localization sub-networks for classification and localization task. And they rely on independent classification loss and localization loss to train the models. This kind of architecture results in the independence between classification and localization task, which hurt the models' localization accuracy. Fitness NMS \cite{tychsen2018fitnessNMS} classifies localization accuracy into 5 levels based on the IoU of regressed boxes and designs sub-networks to predict the probabilities of each localization level independent or dependent of classes for every detection. Then fitness is computed based on these probabilities and combined with the classification score to compute the final detection score, which enhances the correlation between classification and localization task. The enhanced detection score is used as the input for NMS, denoted as Fitness NMS. Similarly, IoU-Net \cite{jiang2018IoUNet} adds an IoU prediction branch parallel with the classification and localization branches to predict the IoU for every detection and the predicted IoU is highly correlated with the localization accuracy. Different from Fitness-NMS, the predicted IoU is directly used as the input for the NMS, denoted as IoU-guided NMS. IoU-aware RetinaNet \cite{2019IoU-aware} attaches an IoU prediction head parallel to the regression head to predict the localization accuracy. During inference, the final confidence is computed by multiplying the predicted IoU and classification score.  MS R-CNN \cite{huang2019MSR-CNN} designs a MaskIoU head to predict the IoU of the predicted masks aiming to solve the problem of the weak correlation between classification score and mask quality. During inference, the predicted mask IoU is multiplied with the classification score as the final mask confidence, which is highly correlated with the mask quality. Unlike IoU-Net, the enhanced mask confidence is only used to rank the predicted masks when computing COCO AP. Different from these methods, IoU-balanced classification loss directly uses the IoU of positive examples to compute weights assigned to positive examples without sacrificing efficiency.

\textbf{Outliers during training localization subnetwork}. Compared with R-CNN \cite{girshick2014r-cnn} and SPPnet \cite{he2015SPPNet}, Fast R-CNN \cite{girshick2015fast} adopts smooth L1 loss to constrain the gradients of outliers as a constant, which prevents gradient explosion. GHM \cite{li2019gradient} analyzes the example imbalance in one-stage detectors in terms of gradient norm distribution. The analysis demonstrates that for the localization subnetwork of a converged model, there are still a large number of outliers and the gradients can be dominated by these outliers during training, which hurts the training process for accurate object localization. Thus GHM-R is proposed to up-weight easy examples and down-weight outliers based on the gradient density of every example. However, gradient density computation is time-consuming and can slow down the training speed. Libra R-CNN \cite{pang2019libraRCNN} claims that the overall gradient of smooth L1 loss is dominated by the outliers when balancing classification and localization task directly. As a result, balanced L1 loss is proposed to increase the gradient of easy examples and keep the gradient of outliers unchanged. Different from these methods, IoU-balanced localization loss computes the weights of every positive example based on their IoU and up-weights examples with high IoU while down-weighting examples with low IoU.

\section{Method}
\label{S:3}
\subsection{Preliminaries}
Loss functions are extremely important for the performance of object detection models. With the development of object detection models, many different kinds of loss functions have been proposed. For most of the popular object detection models such as Faster R-CNN\cite{ren2015faster}, RetinaNet\cite{lin2017focal} and SSD\cite{liu2016ssd}, cross-entropy loss as Equ.\ref{eq:CE} shows is commonly adopted as the classification loss and smooth L1 loss as Equ.\ref{eq:smoothL1} shows is commonly adopted as the localization loss. $p_i$ and $\hat{p}_i$ represent the predicted classification score and the corresponding ground truth label respectively. For positive examples and negative examples, $\hat{p}_i$ equals to 1 and 0 respectively. $x_i$ equals to $l_i-\hat{g}_i$ where $l_i$ and $\hat{g}_i$ represent the parameterized coordinate vectors of the predicted box and the corresponding ground truth box respectively.

\begin{equation}
    \label{eq:CE}
    \operatorname{CE}({{p}_{i}},{{\hat{p}}_{i}})=-{{\hat{p}}_{i}}\log {{p}_{i}}-(1-{{\hat{p}}_{i}})\log (1-{{p}_{i}})
\end{equation}

\begin{equation}
    \label{eq:smoothL1}
    {{\operatorname{smooth}}_{L1}}(x_i)=\left\{ \begin{matrix}
   \frac{{{x_i}^{2}}}{2\delta } & if\left| x_i \right|\le \delta   \\
   \left| x_i \right|-\frac{\delta }{2} & otherwise  \\
   \end{matrix} \right.
\end{equation}

As the standard cross-entropy loss assigns equal weight(1) for all the positive examples with different localization accuracy, it will drive the models to learn as high classification scores as possible for all the positive examples regardless of their localization accuracy. As a result, the classification score will have low correlation with the localization accuracy. For the localization loss, because the number of positive examples with low localization accuracy is larger and the gradient of these kind of positive examples is larger, the gradient produced by these examples will dominate the training process of the localization branch, which hurts the localization accuracy of models. So we propose IoU-balanced loss functions to make the positive examples to adaptively adjust their weight based on their localization accuracy. Both these losses can make object detection models more powerful for accurate localization. These two losses will be introduced in details in the following sub-sections.

\subsection{IoU-balanced Classification Loss}
\label{subS:3.1}
As demonstrated above, the weak correlation between classification and localization tasks will hurt the models' performance during NMS and computing COCO AP. Thus IoU-balanced classification loss is proposed to enhance the correlation between the classification and localization task as Equ.\ref{eq1},\ref{eq2} show.

\begin{equation}
\label{eq1}
{{L}_{cls}}=\sum\limits_{i\in Pos}^{N}{{{w}_{i}}(iou_i)*\operatorname{CE}({{p}_{i}},{{{\hat{p}}}_{i}})+\sum\limits_{i\in Neg}^{M}{\operatorname{CE}({{p}_{i}},{{{\hat{p}}}_{i}})}}
\end{equation}

\begin{equation}
\label{eq2}
{{w}_{i}}(io{{u}_{i}})=iou_{i}^{\eta}\frac{\sum\limits_{i=1}^{N}{\operatorname{CE}({{p}_{i}},{{{\hat{p}}}_{i}})}}{\sum\limits_{i=1}^{N}{iou_{i}^{\eta}\operatorname{CE}({{p}_{i}},{{{\hat{p}}}_{i}})}}
\end{equation}

$Pos$ and $Neg$ represent the sets of positive training examples and negative training examples respectively. ${iou}_i$ represents the regressed IoU for each regressed positive example. The weights $w_i(iou_i)$ assigned to positive examples are positively correlated with the IoU between the regressed bounding boxes and their corresponding ground truth boxes. As a result, the examples with high IoU are up-weighted and the ones with low IoU are down-weighted adaptively based on their IoU after bounding box regression. During training, the examples with higher IoU will contribute larger gradients and thus the model is easier to learn higher classification scores for these examples. On the contrary, the gradients contributed by examples with low IoU will be suppressed and thus the trained models are more likely to learn lower classification scores for these examples. In this way, the correlation between classification scores and localization accuracy is enhanced as demonstrated by Figure \ref{fig:score_detections1} in the following experiment. The parameter $\eta$ is used to control to what extent the IoU-balanced classification loss focuses on examples with high IoU and suppresses examples with low IoU. Besides, the normalization strategy as Equ.\ref{eq2} shows is adopted to keep the sum of classification loss for positive examples unchanged compared with the standard cross-entropy loss during training.

\subsection{IoU-balanced Localization Loss}
\label{subS:3.2}
As analyzed above, if the training process is dominated by the gradients of outliers, the localization accuracy of detectors will get hurt. Thus, we propose IoU-balanced localization loss to up-weight the examples with high IoU and down-weight the examples with low IoU as Equ.\ref{eq3},\ref{eq4} show.

\begin{equation}
\label{eq3}
    {{L}_{loc}}=\sum\limits_{i\in Pos}^{N}{\sum\limits_{m\in cx,cy,w,h}{{{w}_{i}}(io{{u}_{i}})*\text{smoot}{{\text{h}}_{L1}}(l_{i}^{m}-\hat{g}_{i}^{m})}}
\end{equation}

\begin{equation}
\label{eq4}
    {{w}_{i}}(io{{u}_{i}})={{w}_{loc}}*iou_{i}^{\lambda }
\end{equation}

\begin{equation}
\label{eq5}
    {{w}_{i}}(io{{u}_{i}})=iou_{i}^{\lambda }*\frac{\sum\limits_{i\in Pos}^{N}{\sum\limits_{m\in cx,cy,w,h}{\text{smooth}_{L1}(l_{i}^{m}-\hat{g}_{i}^{m})}}}{\sum\limits_{i\in Pos}^{N}{\sum\limits_{m\in cx,cy,w,h}{iou_{i}^{\lambda }*{\text{smooth}}_{L1}(l_{i}^{m}-\hat{g}_{i}^{m})}}}
\end{equation}

$(l_i^{cx}, l_i^{cy}, l_i^w, l_i^h)$ represents the parameterized coordinates of the predicted box and $(\hat{g}_i^{cx}, \hat{g}_i^{cy}, \hat{g}_i^w, \hat{g}_i^h)$ represents the parameterized coordinates of the corresponding ground truth box. The parameterization strategy is the same as that in R-CNN \cite{girshick2014r-cnn}. The parameter $\lambda$ is designed to control to what extent IoU-balanced localization loss focuses on inliers and suppresses outliers. The localization loss weight $w_{loc}$ is manually adjusted to keep the sum of localization loss unchanged compared with the original smooth L1 loss for the first iteration of the training procedure. Normalization strategy can also be used to keep the sum of localization loss unchanged during the whole training procedure as Equ.\ref{eq5} shows. However, the experiments show that this normalization strategy is slightly inferior compared with manually adjusting $w_{loc}$. This may be caused by that the normalization factor is decreased as the IoUs of positive examples increase during training. Thus, the strategy of manually adjusting $w_{loc}$ is adopted in all the following experiments.

We constrains that the gradients are not propagated from  ${{w}_{i}}(iou_i) $ to $l_{i}^{m}$ . Denoting $d=l^{m}-\hat{g}^{m}$ , the gradient of IoU-balanced smooth L1 loss with respect to  $l^m$ can be expressed as:

\begin{equation}
\label{eq6}
\frac{\partial w(iou)*\text{smooth}_{L1}}{\partial {{l}^{m}}}=\left\{
\begin{matrix}
w(iou)*\frac{d}{\delta}   &  {if\quad \left| d \right|\le \delta,}\\
w(iou)*sign(d)            &  {otherwise.}
\end{matrix} \right.
\end{equation}

The IoU function representing the relationship between IoU and $d$ is complex and Bounded IoU \cite{tychsen2018fitnessNMS} simplifies this function by computing an upper bound of the IoU function. The same idea is adopted in this paper and readers can refer to Bounded IoU for more details. Given an anchor or a proposal ${{b}_{s}}=({{x}_{s}},{{y}_{s}},{{w}_{s}},{{h}_{s}})$, an associated ground truth box ${{b}_{t}}=({{x}_{t}},{{y}_{t}},{{w}_{t}},{{h}_{t}})$ and a predicted bounding box ${{b}_{p}}=({{x}_{p}},{{y}_{p}},{{w}_{p}},{{h}_{p}})$ , the upper bound of the IoU function is as follows:

\begin{equation}
    \label{eq7}
    	\text{io}{{\text{u}}_{B}}(x\text{,}{{b}_{t}})=\frac{{{w}_{t}}-\left| \Delta x \right|}{{{w}_{t}}+\left| \Delta x \right|}
\end{equation}

\begin{equation}
    \label{eq8}
    \text{io}{{\text{u}}_{\text{B}}}(w,{{b}_{t}})=\min (\frac{{{w}_{p}}}{{{w}_{t}}},\frac{{{w}_{t}}}{{{w}_{p}}})
\end{equation}
where $\Delta{x} ={{x}_{p}}-{{x}_{t}}$. Because there exists that $\left| {{d}^{cx}} \right|=\left| \Delta x/{{w}_{s}} \right|$, $\left| {{d}^{w}} \right|=\left| \log ({{w}_{p}}/{{w}_{t}}) \right|$, we can get:

\begin{equation}
    \label{eq9}
    	\text{io}{{\text{u}}_{\text{B}}}(x\text{,}{{b}_{t}})=\frac{{{w}_{t}}-{{w}_{s}}\left| {{d}^{cx}} \right|}{{{w}_{t}}+{{w}_{s}}\left| {{d}^{cx}} \right|}
\end{equation}

\begin{equation}
    \label{eq10}
    	{{\operatorname{iou}}_{\text{B}}}(w,{{b}_{t}})={{e}^{-\left| {{d}^{w}} \right|}}
\end{equation}
which satisfies $\left| {{d}^{cx}} \right|\le {{w}_{t}}/{{w}_{s}}$ to ensure ${{\operatorname{iou}}_{B}}(x,{{b}_{t}})\ge 0$. $\text{io}{{\text{u}}_{\text{B}}}(y,{{b}_{t}})$ and $\text{io}{{\text{u}}_{\text{B}}}(h,{{b}_{t}})$ are similar to ${{\operatorname{iou}}_{\text{B}}}(x\text{,}{{b}_{t}})$ and ${{\operatorname{iou}}_{\text{B}}}(w,{{b}_{t}})$ respectively. Assuming that $\delta =0.111$ and ${{w}_{t}}={{w}_{s}}$, we have
\begin{equation}
    \label{eq11}
    \frac{\partial w(io{{u}_{B}})*{{\operatorname{smooth}}_{L1}}}{\partial d}=\left\{ \begin{matrix}
   {{w}_{loc}}*{{\left( \frac{1-\left| d \right|}{1+\left| d \right|} \right)}^{\lambda }}\frac{d}{\delta } & if\left| d \right|\le \delta   \\
   {{w}_{loc}}*{{\left( \frac{1-\left| d \right|}{1+\left| d \right|} \right)}^{\lambda }}\operatorname{sign}(d) & otherwise  \\
\end{matrix} \right.
\end{equation}
for $d={{d}^{cx}}$ or $d={{d}^{cy}}$ and
\begin{equation}
    \label{eq12}
    \frac{\partial w(io{{u}_{B}})*{{\operatorname{smooth}}_{L1}}}{\partial d}=\left\{ \begin{matrix}
   {{w}_{loc}}*{{e}^{-\lambda \left| d \right|}}\frac{d}{\delta } & if\left| d \right|\le \delta   \\
   {{w}_{loc}}*{{e}^{-\lambda \left| d \right|}}\operatorname{sign}(d) & otherwise  \\
\end{matrix} \right.
\end{equation}
for $d={{d}^{w}}$ or $d={{d}^{h}}$.

\begin{figure}[h]
\centering\includegraphics[width=0.48\linewidth]{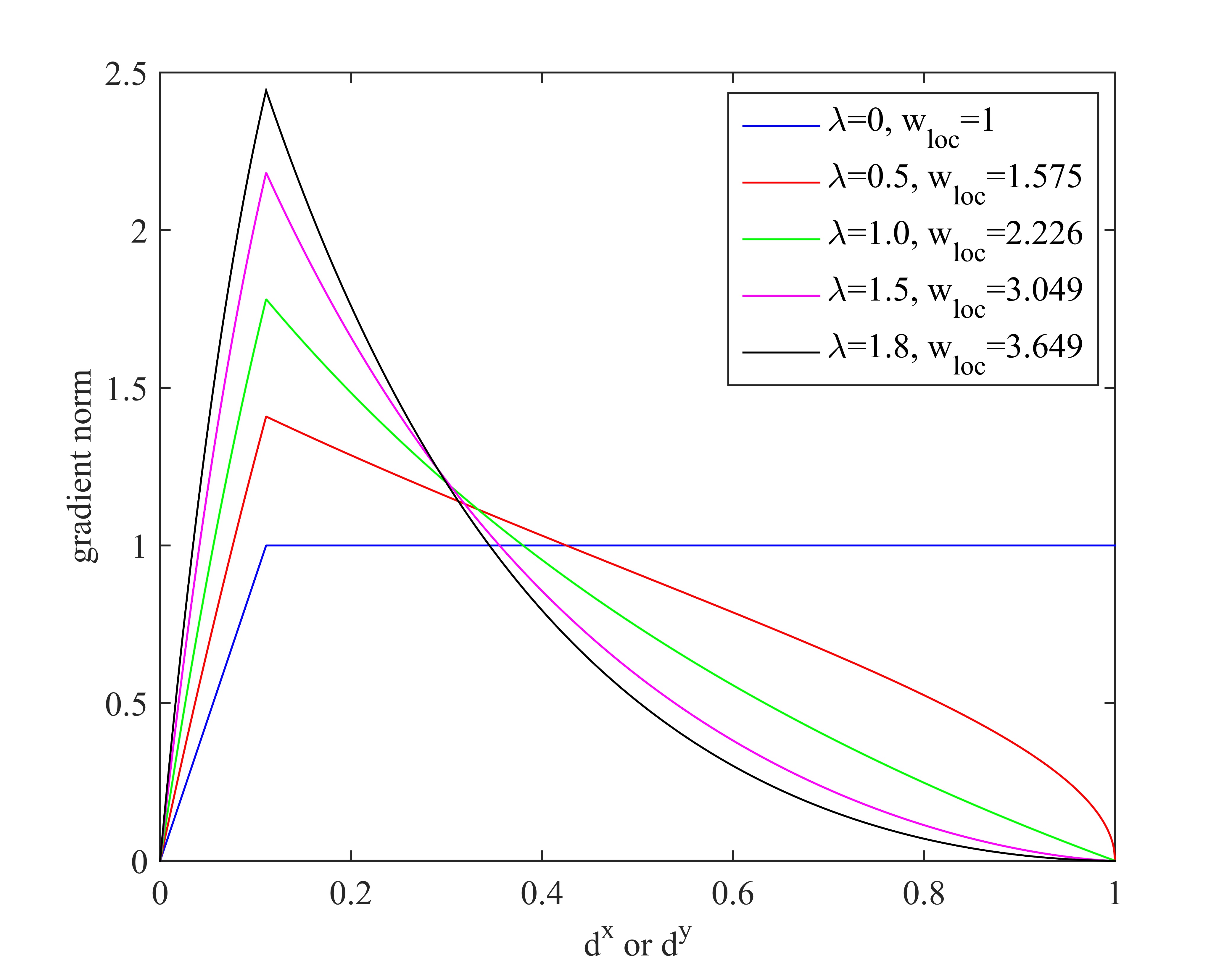}
\centering\includegraphics[width=0.48\linewidth]{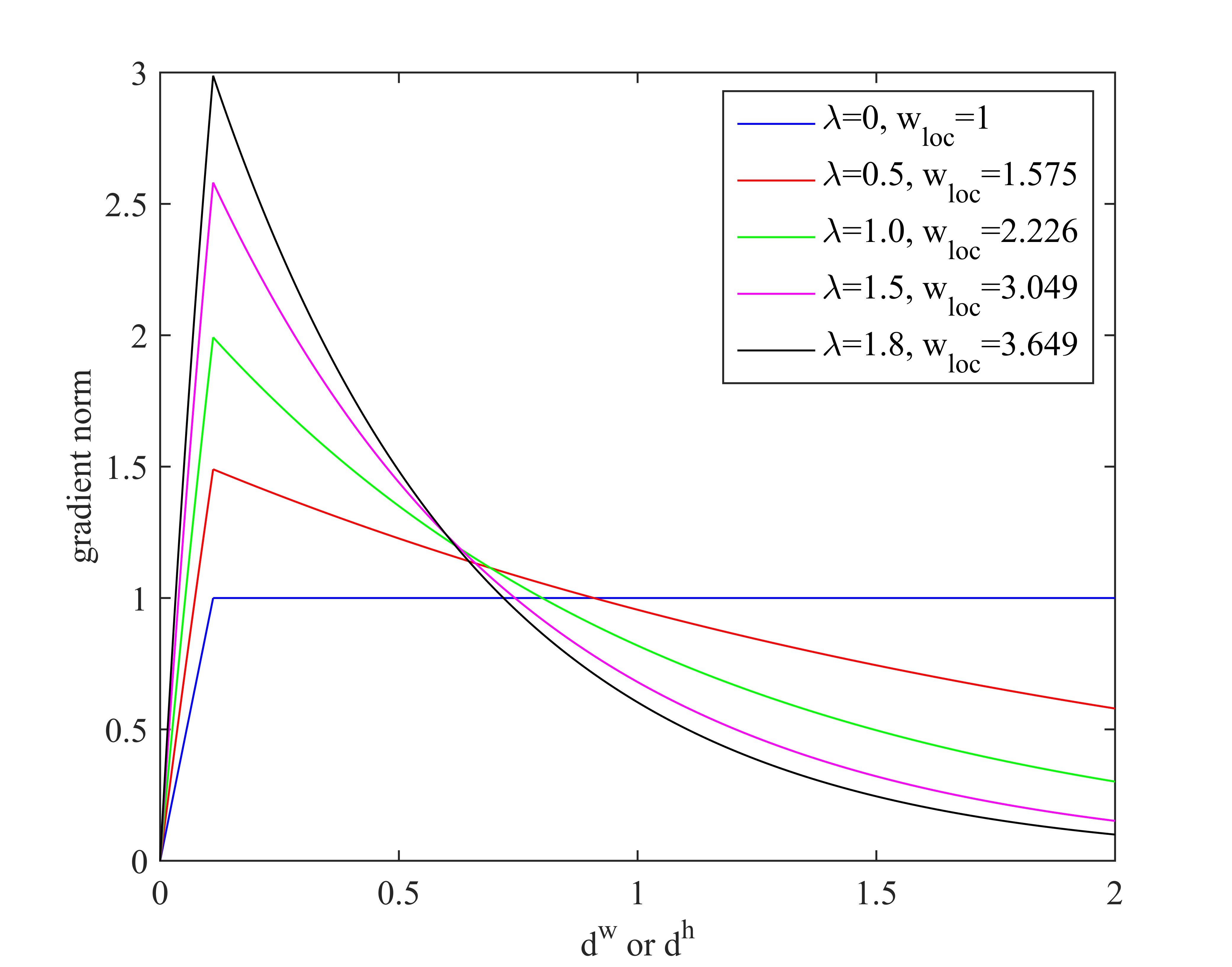}
\caption{The gradient norm of the standard smooth L1 loss ($\lambda =0$) and the upper bound of gradient norm for IoU-balanced smooth L1 loss $(\lambda =0.5,1.0,1.5,1.8)$ with respect to ${{d}^{cx}}$, ${{d}^{cy}}$, ${{d}^{w}}$, ${{d}^{h}}$. The localization weight ${{w}_{loc}}$ is manually adjusted to keep the sum of localization loss unchanged when $\lambda $ is changed compared with the standard smooth L1 loss.}
\label{fig:gradient_norm}
\end{figure}

The gradient norm of standard smooth L1 loss ($\lambda =0$) and the upper bound of gradient norm for IoU-balanced smooth L1 loss $(\lambda =0.5,1.0,1.5,1.8)$ with respective to ${{d}^{cx}}$, ${{d}^{cy}}$, ${{d}^{w}}$, ${{d}^{h}}$ are visualized in Figure \ref{fig:gradient_norm}. Compared with the standard smooth L1 loss, IoU-balanced smooth L1 loss can increase the gradient norm of inliers and reduce the gradient norm of outliers, making the model more powerful for accurate localization.

\section{Experiments}
\label{S:4}

\subsection{Experimental Settings}
We evaluate the proposed IoU-balanced losses on the popular single-stage object detection models including anchor-based detectors(SSD\cite{liu2016ssd}, RetinaNet\cite{lin2017focal}) and anchor-free detector FoveaBox\cite{kong2020foveabox}. Besides, we also analyze the effectiveness of IoU-balanced losses on the two-stage detector Faster R-CNN. And only the loss functions in these models are changed during training for a fair comparison.

\textbf{Dataset}. We evaluate our method on three popular object detection datasets including MS COCO\cite{lin2014microCOCO}, PASCAL VOC\cite{everingham2010pascalVOC} and Cityscapes\cite{Cordts2016Cityscapes}. For MS COCO, it consists of 118k images for training (\textit{train-2017}), 5k images for validation (\textit{val-2017}) and 20k images with no disclosed labels for test (\textit{test-dev}). There are totally over 500k annotated object instances from 80 categories in the dataset. For PASCAL VOC, the VOC2007 contains 5011 images for training (\textit{VOC2007 trainval}) and 4952 for test (\textit{VOC2007 test}). The VOC2012 contains 17125 images for training (\textit{VOC2012 trainval}) and 5138 for test (\textit{VOC2012 test}). We train models on the union of \textit{VOC2007 trainval} and \textit{VOC2012 trainval} and evaluate models on \textit{VOC2007 test}. For Cityscapes, it consists of a large, diverse set of stereo video sequences recorded in streets from 50 different cities. 5000 of these images have high-quality pixel-level annotations and 20000 additional images have coarse annotations. For the fine annotated images, it's split into 2975 images for training, 500 images for validation and 1525 images with no annotations for test. In our experiments, only fine annotated images are used. We train models on the training split and evaluate models on the validation split.

\textbf{Evaluation Metrics}. For experimental results on MS COCO and Citys- capes datasets, the standard COCO-style metrics are adopted which includes AP (averaged on IoUs from 0.5 to 0.95 with an interval of 0.05), $\text{A}{{\text{P}}_{50}}$ (AP at IoU threshold 0.5), $\text{A}{{\text{P}}_{75}}$ (AP at IoU threshold 0.75), $\text{A}{{\text{P}}_{S}}$ (AP for small scales), $\text{A}{{\text{P}}_{M}}$ (AP for medium scales) and $\text{A}{{\text{P}}_{L}}$ (AP for large scales). For experimental results on PASCAL VOC, we report AP at different IoU thresholds and the averaged AP.

\textbf{Implementation Details}. All the experiments are implemented based on PyTorch and MMDetection \cite{chen2019mmdetection}. As only 2 GPUs are available, linear scaling rule \cite{goyal2017accurate} is adopted to adjust the learning rate during training. For the main results, all the models are evaluated on COCO \textit{test-dev}. Except for SSD, all the IoU-balanced models and the baselines are trained for a total of 12 epochs using image scale of [800, 1333]. IoU-balanced SSDs and their baselines are trained for a total of 120 epochs with image scale of [300, 300] and [512, 512]. Some papers report the main results obtained by training the models for totally 1.5 longer times and with scale jitter. These tricks are not adopted in our experiments. In the ablation studies, RetinaNet with ResNet50 as backbone are trained on \textit{train-2017} and evaluated on \textit{val-2017} using image scale of [600, 1000]. Faster R-CNN with backbone ResNet50 is trained on \textit{train-2017} and evaluated on \textit{val-2017} using image scale of [600, 1000]. For the experiments on PASCAL VOC and Cityscapes, the best parameters for IoU-balanced losses searched in the COCO experiments are adopted. For Cityscapes, we pretrain models on the COCO and finetune on the Cityscapes. If not specified, all the other settings are kept the same as the default settings provided by MMDdetection.

\begin{table}[h]
\centering
\resizebox{1\linewidth}{!}{

\begin{tabular}{l l l l l l l l l}
\hline
{Model} & {Backbone} & {Schedule} & {$\text{AP}$} & {$\text{AP}_{50}$} & {$\text{AP}_{75}$} & {$\text{AP}_S$} & {$\text{AP}_M$} & {$\text{AP}_L$}\\
\hline
YOLOv2 \cite{redmon2017yolo9000} & DarkNet-19 & - & 21.6 & 44.0 & 19.2 & 5.0  & 22.4 & 35.5 \\
YOLOv3 \cite{redmon2018yolov3}   & DarkNet-53 & - & 33.0 & 57.9 & 34.4 & 18.3 & 35.4 & 41.9 \\
SSD300 \cite{liu2016ssd}         & VGG16      & - & 23.2 & 41.2 & 23.4 & 5.3  & 23.2 & 39.6 \\
SSD512 \cite{liu2016ssd}         & VGG16      & - & 26.8 & 46.5 & 27.8 & 9.0  & 28.9 & 41.9 \\
Faster R-CNN \cite{ren2015faster}& ResNet-101-FPN & - & 36.2 & 59.1 & 39.0 & 18.2 & 39.0 & 48.2 \\
Deformable R-FCN \cite{dai2017deformableConv} & Inception-ResNet-v2 & - & 37.5 & 58.0 & 40.8 & 19.4 & 40.1 & 52.5 \\
Mask R-CNN \cite{he2017mask} & ResNet-101-FPN & - & 38.2 & 60.3 & 41.7 & 20.1 & 41.1 & 50.2 \\
\hline
Faster R-CNN* & ResNet-50-FPN & 1x & 36.2 & 58.5 & 38.9 & 21.0 & 38.9 & 45.3 \\
Faster R-CNN* & ResNet-101-FPN & 1x & 38.8 & 60.9 & 42.1 & 22.6 & 42.4 & 48.5 \\
FoveaBox* & ResNet-50-FPN & 1x & 37.0 & 56.7 & 39.1 & 20.3 & 40.0 & 45.6 \\
SSD300* & VGG16 & 120e & 25.7 & 44.2 & 26.4 & 7.0 & 27.1 & 41.5 \\
SSD512* & VGG16 & 120e & 29.6 & 49.5 & 31.2 & 11.7 & 33.0 & 44.2 \\
RetinaNet* & ResNet-50-FPN & 1x & 35.9 & 55.8 & 38.4 & 19.9 & 38.8 & 45.0 \\
RetinaNet* & ResNet-101-FPN & 1x & 38.1 & 58.5 & 40.8 & 21.2 & 41.5 & 48.2 \\
RetinaNet* & ResNeXt-32x4d-101-FPN & 1x & 39.4 & 60.2 & 42.3 & 22.5 & 42.8 & 49.8 \\
\hline
IoU-balanced FoveaBox & ResNet-50-FPN & 1x & 38.0 & 56.9 & 40.1 & 21.2 & 40.8 & 46.7 \\
IoU-balanced SSD300 & VGG16 & 120e & 27.4 & 45.0 & 28.8 & 8.5 & 28.9 & 43.0 \\
IoU-balanced SSD512 & VGG16 & 120e & 30.9 & 50.1 & 32.9 & 12.6 & 34.5 & 45.1 \\
IoU-balanced RetinaNet & ResNet-50-FPN & 1x & 37.0 & 56.2 & 39.7 & 20.6 & 39.8 & 46.3 \\
IoU-balanced RetinaNet & ResNet-101-FPN & 1x & 39.2 & 58.7 & 42.3 & 21.5 & 42.4 & 49.4 \\
IoU-balanced RetinaNet & ResNeXt-32x4d-101-FPN & 1x & 40.5 & 60.3 & 43.6 & 23.0 & 43.7 & 51.0 \\
\hline
\end{tabular}
}
\caption{ Comparison with the state-of-the-art methods on COCO \textit{test-dev}. The symbol "*" means the reimplementation results in MMDetection \cite{chen2019mmdetection}. The training schedule is the same as Detectron \cite{Detectron2018}."1x" and "120e" means the model is trained for 12 epochs and 120 epochs respectively.}
\label{table:1}
\end{table}

\subsection{Main Results}
In the main results, the performance of our proposed method is compared with the state-of-the-art object detection models on the COCO \textit{test-dev} in Table \ref{table:1}. For a fair comparison, we adopt the reimplemented models in MMDetectioin \cite{chen2019mmdetection} as the baselines. For anchor-based detectors, IoU-balanced loss functions can consistently improve AP by $1.1\%$ for RetineNet with different backbones and largely improve AP by $1.7\%$, $1.3\%$ for SSD300 and SSD512 respectively. For anchor-free detector FoveaBox, IoU-balanced losses improve the AP by $1.0\%$. This demonstrates that IoU-balanced losses are effective on different kinds of single-stage detectors, even on the state-of-the-art anchor-free detector. In addition, the improvement for $\text{A}{{\text{P}}_{50}}$ is only $0.1\%\sim0.8\%$ while that for $\text{AP}_{75}$ is $1.0\%\sim2.4\%$ which demonstrates the effectiveness of IoU-balanced loss functions for accurate localization. Compared with two-stage detector Faster R-CNN, IoU-balanced RetinaNets with the same backbone have surpassed Faster R-CNN by $0.4\%\sim0.8\%$ on AP and by $0.2\%\sim0.8\%$ even on $\text{AP}_{75}$.

\begin{table}[h]
\centering
\resizebox{1\linewidth}{!}{
\begin{tabular}{l l l l l l l l l l l l}
\hline
{IoU-Cls} & {IoU-Loc} & {$\text{AP}$} & {$\text{AP}_{50}$} & {$\text{AP}_{75}$} & {$\text{AP}_S$} & {$\text{AP}_M$} & {$\text{AP}_L$} & {$\text{AP}_{60}$} & {$\text{AP}_{70}$} & {$\text{AP}_{80}$} & {$\text{AP}_{90}$}\\
\hline
        &         & 34.4 & 53.9 & 36.6 & 17.2 & 38.2 & 48.0 & 49.2 & 41.9 & 30.0 & 11.2\\
$\surd$ &         & 35.1 & 54.6 & 37.5 & 18.4 & 38.5 & 47.8 & 50.2 & 42.7 & 31.0 & 11.4\\
        & $\surd$ & 35.2 & 53.7 & 37.6 & 17.9 & 39.3 & 48.5 & 49.3 & 42.3 & 31.8 & 13.1\\
$\surd$ & $\surd$ & 35.7 & 54.3 & 38.0 & 17.7 & 39.4 & 48.8 & 50.0 & 43.0 & 32.1 & 13.5\\
\hline
\end{tabular}
}
\caption{ The effectiveness of IoU-balanced classification loss and IoU-balanced localization loss for RetinaNet-ResNet50 on COCO \textit{val-2017}. The abbreviations of "IoU-Cls" and "IoU-Loc" represent IoU-balanced classification loss and IoU-balanced localization loss respectively.}
\label{table:ablation-retinanet}
\end{table}

\subsection{Analysis}
\textbf{Component Analysis}. The effectiveness of different components is analyzed as Table \ref{table:ablation-retinanet} shows. IoU-balanced classification loss and IoU-balanced localization loss can improve AP by $0.7\%$ and $0.8\%$ respectively and combining them can improve AP by $1.3\%$. In addition, IoU-balanced classification loss has consistent improvement for average precision at different IoU threshold. This demonstrates the importance of enhancing the correlation between classification and localization tasks. IoU-balanced localization loss slightly decreases $AP_{50}$  by $0.2\%$ but substantially improves $AP_{80}$  and $AP_{90}$  by $1.8\%$ and $1.9\%$ respectively. This demonstrates that up-weighting the gradients of inliers while down-weighting the gradients of outliers for the localization loss is especially beneficial for accurate localization.

\begin{table}[h]
\centering
\begin{tabular}{l l l l l}
\hline
$\eta$ & {$\text{AP}$} & $\lambda$ & $w_{loc}$ & $\text{AP}$ \\
\hline
0   & 34.4 & 0   & 1.0   & 34.4 \\
1.0 & 34.7 & 0.5 & 1.575 & 35.0 \\
1.4 & 35.0 & 1.0 & 2.226 & 35.1 \\
1.5 & 35.1 & 1.5 & 3.049 & 35.2 \\
1.6 & 34.9 & 1.8 & 3.649 & 35.1 \\
\hline
\end{tabular}
\caption{ The effectiveness of varying $\eta $ in IoU-balanced classification loss and $\lambda $ in IoU-balanced localization loss respectively.}
\label{table:4}
\end{table}

\begin{figure}[h]
\centering
\subfloat[]{\label{fig:score_detections1}
\includegraphics[height=0.36\linewidth]{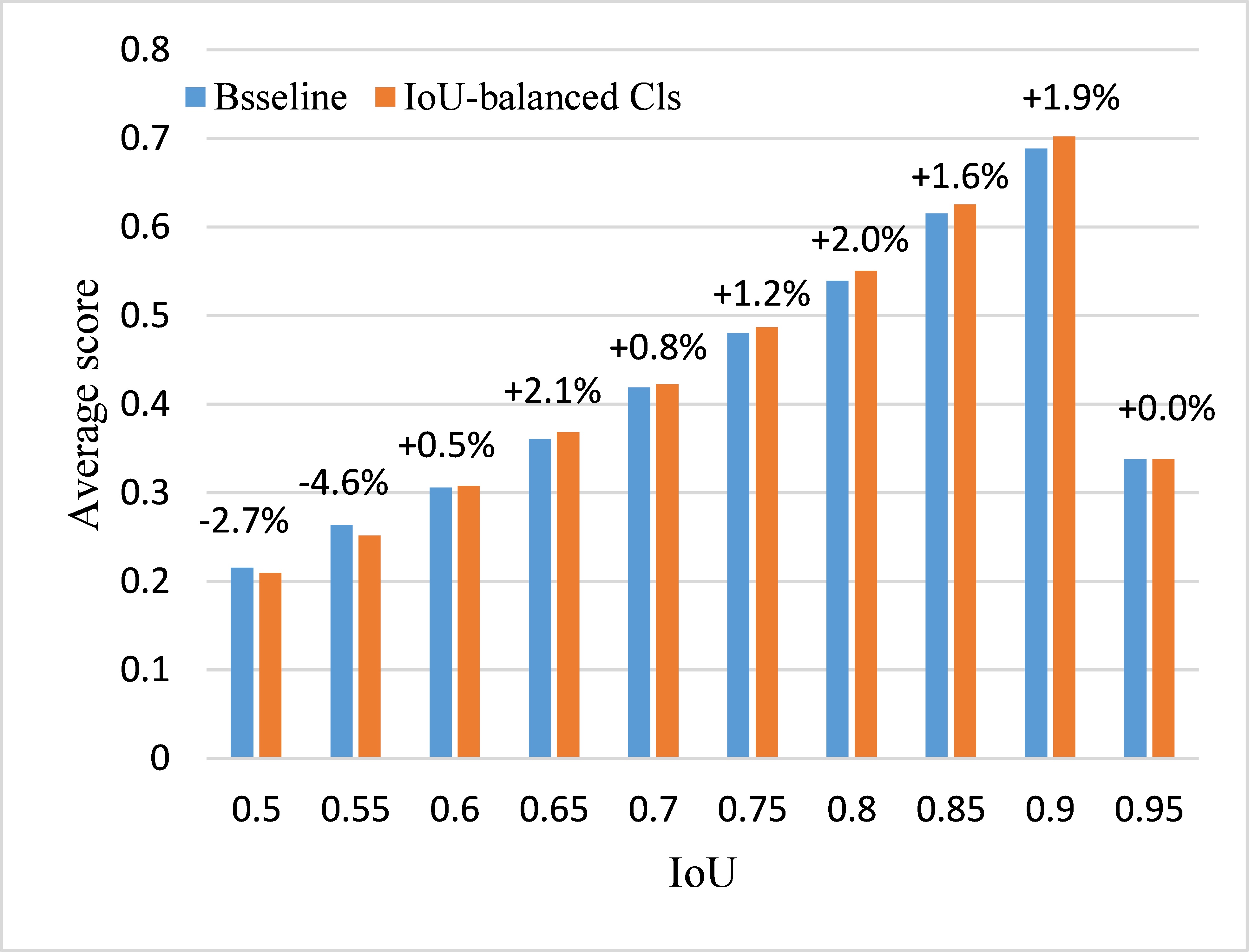}}
\subfloat[]{\label{fig:score_detections2}
\includegraphics[height=0.36\linewidth]{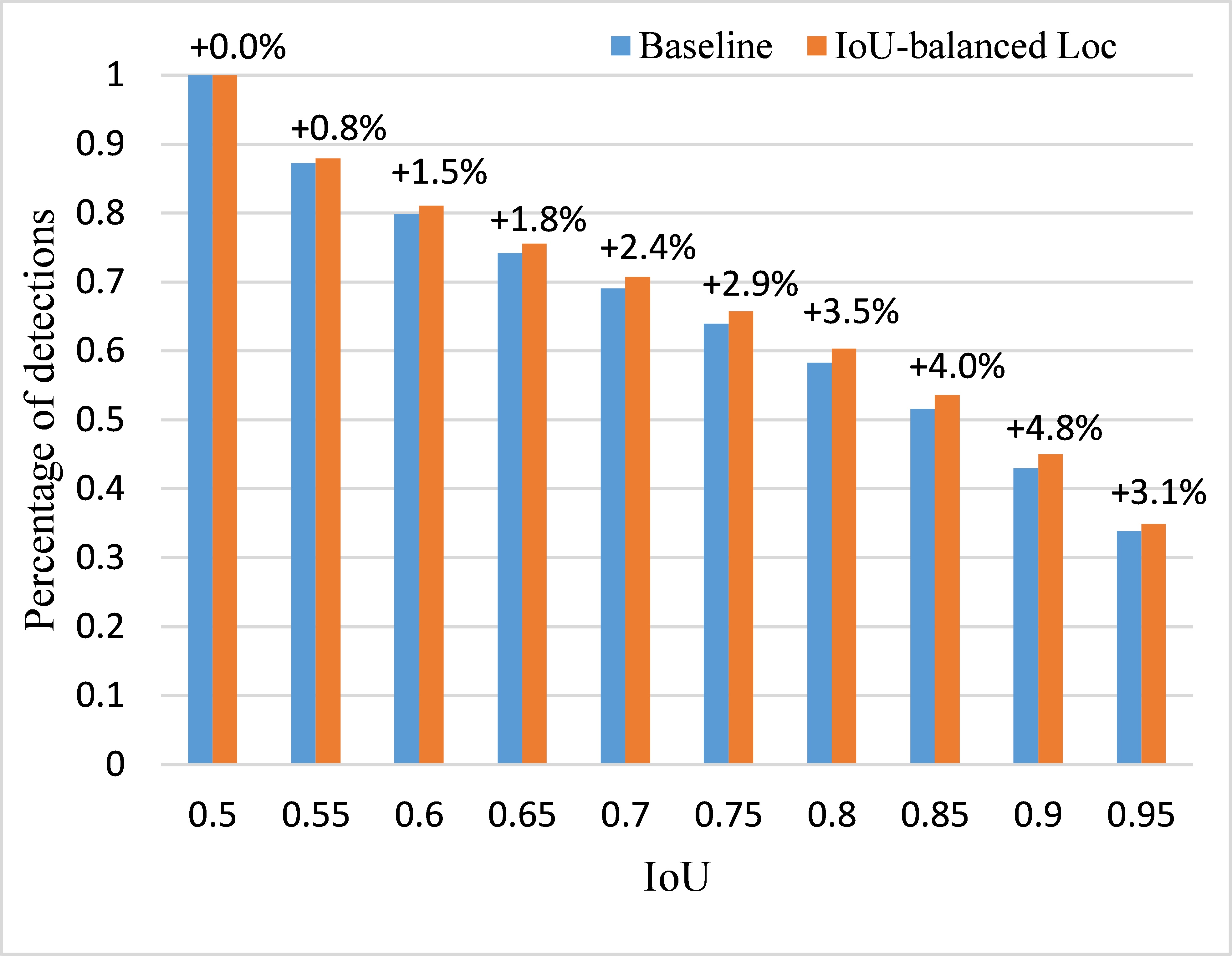}}
\caption{(a) Average classification scores of detections within different IoU range. IoU-balanced classification loss can increases the average classification scores for the detections with high IoU($0.6\sim0.95$) by $0.5\%\sim2.1\%$ while decreasing the average classification scores for the detections with low IoU($0.5\sim0.55$) by $2.7\%\sim4.6\%$. (b) The percentage of detections that have higher IoU than the corresponding IoU thresholds. IoU-balanced localization loss can increase the percentage of detections with high IoU by $0.8\%\sim4.8\%$ relative to the baseline.}
\label{fig:score_detections}
\end{figure}

\textbf{Ablation Studies on IoU-balanced Classification Loss}. The parameter $\eta $ in IoU-balanced classification loss controls to what extent the model focuses on the positive examples with high IoU. As Table \ref{table:4} shows, the model can achieve the best performance of AP $35.1\%$ when $\eta $ equals to 1.5. As shown in Figure \ref{fig:score_detections1}, compared with the baseline, IoU-balanced classification loss can increase the average classification scores for the examples with high IoU by $0.5\%\sim2.1\%$ and decrease the average classification scores for the examples with low IoU by $2.7\%\sim4.6\%$, which demonstrates that the correlation between classification and localization task is enhanced by the IoU-balanced classification loss.

\textbf{Ablation Studies on IoU-balanced Localization Loss}. As Figure \ref{fig:gradient_norm} shows, the parameter $\lambda $ in IoU-balanced localization loss controls to which extent the model increases the gradient norm of inliers and decreases the gradient norm of outliers. The localization loss weight ${{w}_{loc}}$ is manually adjusted to keep the sum of localization loss unchanged when changing the parameter $\lambda $. As Table \ref{table:4} shows, the best performance of $\text{AP}$ $35.2\%$ is obtained when $\lambda $ equals to 1.5. As shown in Figure \ref{fig:score_detections2}, IoU-balanced localization loss increases the percentage of detections with high IoU by $0.8\%\sim4.8\%$ relative to the baseline model. This demonstrates that the IoU-balanced localization loss can make the model more powerful for accurate localization.

\begin{table}[h]
\centering
\resizebox{1\linewidth}{!}{
\begin{tabular}{l l l l l l l l l l l l}
\hline
{method} & {Backbone} & {$\text{AP}$} & {$\text{AP}_{50}$} & {$\text{AP}_{75}$} & {$\text{AP}_S$} & {$\text{AP}_M$} & {$\text{AP}_L$} & {$\text{AP}_{60}$} & {$\text{AP}_{70}$} & {$\text{AP}_{80}$} & {$\text{AP}_{90}$}\\
\hline
         & ResNet-18         & 30.8 & 49.6 & 32.4 & 16.1 & 34.0 & 40.7 & 45.0 & 37.6 & 26.1 & 8.6\\
         & ResNet-50         & 35.6 & 55.5 & 38.3 & 20.0 & 39.6 & 46.8 & 51.0 & 43.2 & 31.1 & 11.3\\
baseline & ResNet-101        & 37.7 & 57.5 & 40.4 & 21.1 & 42.2 & 49.5 & 53.3 & 46.0 & 33.7 & 13.0\\
         & ResNeXt-32x4d-101 & 39.0 & 59.4 & 41.7 & 22.6 & 43.4 & 50.9 & 55.2 & 47.6 & 34.9 & 14.1\\
\hline
                    & ResNet-18         & 32.0 & 49.7 & 34.0 & 16.3 & 34.8 & 43.3 & 45.8 & 39.0 & 28.0 & 10.6\\
                    & ResNet-50         & 36.7 & 55.7 & 39.3 & 20.7 & 40.6 & 48.0 & 51.5 & 44.3 & 32.8 & 13.9\\
IoU-balanced losses & ResNet-101        & 38.8 & 58.0 & 41.6 & 21.1 & 43.1 & 51.3 & 54.1 & 46.9 & 35.3 & 15.4\\
                    & ResNeXt-32x4d-101 & 40.4 & 60.2 & 43.1 & 23.2 & 44.8 & 52.2 & 55.9 & 48.6 & 36.9 & 16.3\\
\hline
\end{tabular}
}
\caption{ The effectiveness of IoU-balanced losses for RetinaNet with different backbones on COCO \textit{val-2017} with the image scale [800, 1333].}
\label{table:ablation-DifferentBackbones}
\end{table}

\textbf{Effectiveness on Different Backbones.} To validate the effectiveness of IoU-balanced losses on different backbones, we conduct experiments across different backbones on RetinaNet with image scale [800, 1333]. As shown in Table \ref{table:ablation-DifferentBackbones}, IoU-balanced losses can consistently improve AP by $1.1\%\sim1.4\%$ for different backbones, which shows that IoU-balanced losses are robust to different backbones. And $\text{AP}_{50}$ and $\text{AP}_{60}$ are improved by $0.1\%\sim0.8\%$ while $\text{AP}_{80}$ and $\text{AP}_{90}$ are improved by $1.6\%\sim2.6\%$. It's obvious that IoU-balaced losses are especially effective for improving the models' localization accuracy regardless of the models' capacity.

\begin{figure}[h]
\centering
\subfloat[]{\label{Fig:visualization1}
\includegraphics[width=0.18\linewidth]{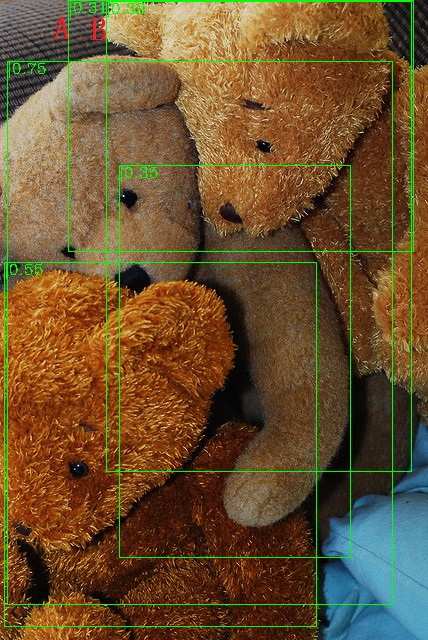}}
\subfloat[]{\label{Fig:visualization2}
\includegraphics[width=0.18\linewidth]{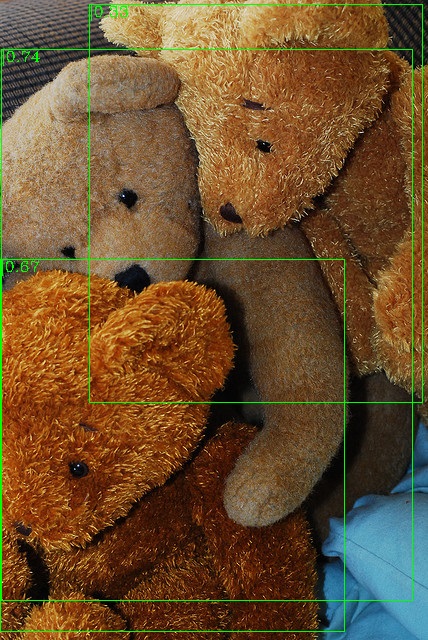}}
\subfloat[]{\label{Fig:visualization3}
\includegraphics[width=0.32\linewidth]{FIG3c.jpg}}
\subfloat[]{\label{Fig:visualization4}
\includegraphics[width=0.32\linewidth]{FIG3d.jpg}}
\caption{Visualization of detection results from the baseline (a,c) and IoU-balanced RetinaNet-RsNet50 (b,d).}
\label{fig:visualization}
\end{figure}

\textbf{Qualitative Evaluation}. Some detection results from the baseline and IoU-balanced RetinaNet-ResNet50 on the COCO \textit{val-2017} images are visualized to give a qualitative evaluation of the advantages of IoU-balanced losses. As shown in Figure \ref{fig:visualization}, IoU-balanced losses can increase the classification scores of detections with high IoU while decreasing the classification scores of detections with low IoU. In addition, the localization accuracy can be improved. For example, the top-right bear in Figure \ref{Fig:visualization1} is detected by two boxes A and B with classification score 0.31 and 0.39 respectively when using the baseline model. After the IoU-balanced RetinaNet is applied, the classification score of A is decreased below the visualization threshold 0.2 and the localization accuracy of B is also improved. For the tie in Figure \ref{Fig:visualization3}, the classification score is 0.25 when using the baseline and is increased to 0.40 when IoU-balanced RetinaNet is used.

\subsection{Generalization to Other Datasets}
To validate the generalization ability to other datasets of our method, we also conduct experiments on PASCAL VOC and CityScapes.

\textbf{PASCAL VOC}. As Table \ref{table:PascalVOC} shows, IoU-balanced losses can substantially improve AP by $1.3\%\sim1.5\%$ for SSD and RetinaNet on PASCAL VOC dataset. The improvement for $\text{A}{{\text{P}}_{50}}$ is $0.1\%\sim0.5\%$ while that for $\text{A}{{\text{P}}_{80}}$ and $\text{A}{{\text{P}}_{90}}$ is $1.6\%\sim3.9\%$, demonstrating the effectiveness of IoU-balanced losses for accurate localization. This is similar to the observations in the experimental results on COCO dataset and demonstrates that IoU-balanced losses have generalization ability to other datasets and can be applied to different application scenes.

\begin{table}[h]
\centering
\resizebox{1\linewidth}{!}{
\begin{tabular}{l l l l l l l l}
\hline
{Model} & {Backbone} & {$\text{AP}$} & {$\text{AP}_{50}$} & {$\text{AP}_{60}$} & {$\text{AP}_{70}$} & {$\text{AP}_{80}$} & {$\text{AP}_{90}$}\\
\hline
SSD300 & VGG16 & 49.3 & 77.3 & 71.8 & 60.8 & 42.4 & 14.0 \\
SSD500 & VGG16 & 51.3 & 80.2 & 75.3 & 64.6 & 44.2 & 12.7 \\
RetinaNet & ResNet-50-FPN & 51.9 & 79.5 & 74.9 & 64.3 & 44.7 & 16.5 \\
RetinaNet & ResNet-101-FPN & 54.7 & 80.8 & 76.7 & 66.6 & 49.5 & 20.3 \\
RetinaNet & ResNeXt-32x4d-101-FPN & 56.2 & 82.0 & 78.3 & 68.2 & 51.6 & 22.0 \\
\hline
IoU-balanced SSD300 & VGG16 & 50.6 & 77.8 & 72.4 & 62.4 & 44.9& 14.7 \\
IoU-balanced SSD500 & VGG16 & 52.8 & 80.6 & 75.8 & 66.1 & 47.0 & 16.1 \\
IoU-balanced RetinaNet & ResNet-50-FPN & 53.3 & 79.8 & 75.1 & 65.4 & 47.6 & 19.4 \\
IoU-balanced RetinaNet & ResNet-101-FPN & 56.0 & 81.0 & 77.0 & 68.0 & 51.1 & 23.1 \\
IoU-balanced RetinaNet & ResNeXt-32x4d-101-FPN & 57.7 & 82.1 & 78.3 & 69.5 & 53.7 & 25.1 \\
\hline
\end{tabular}
}
\caption{Experimental Results on PASCAL VOC. The models are trained on the union of \textit{VOC2007 trainval} and \textit{VOC2012 trainval} and evaluated on \textit{VOC2007 test} with image scale [600, 1000]. Default settings are adopted as in MMDetection. IoU-balanced losses adopt the best parameters searched in the COCO experiments.}
\label{table:PascalVOC}
\end{table}

\textbf{Cityscapes}. RetinaNet with ResNet50 and ResNet101 are trained on Cityscapes-train and evaluated on Cityscapes-val with image scale [2048, 800] and [2048, 1024]. As shown in Table \ref{table:CityScapes}, IoU-balanced losses can improve RetinaNet with different backbones by $1.0\%\sim1.2\%$ for AP, which shows that IoU-balanced losses are also effective on the extremely difficult dataset and can be applied to extremely challenging real scenarios such as autonomous driving.

\begin{table}[h]
\centering
\resizebox{1\linewidth}{!}{
\begin{tabular}{l l l l l l l l}
\hline
{Model} & {Backbone} & {$\text{AP}$} & {$\text{AP}_{50}$} & {$\text{AP}_{75}$} & {$\text{AP}_{S}$} & {$\text{AP}_{M}$} & {$\text{AP}_{L}$}\\
\hline
RetinaNet & ResNet-50-FPN & 39.1 & 64.1 & 38.5 & 15.7 & 40.0 & 59.1 \\
RetinaNet & ResNet-101-FPN & 39.7 & 64.6 & 40.6 & 15.1 & 40.3 & 60.5 \\
\hline
IoU-balanced RetinaNet & ResNet-50-FPN & 40.3 & 64.6 & 41.3 & 17.6 & 40.9 & 60.3 \\
IoU-balanced RetinaNet & ResNet-101-FPN & 40.7 & 65.2 & 41.8 & 16.0 & 41.1 & 62.3 \\
\hline
\end{tabular}
}
\caption{Experimental Results on Cityscapes. The models are trained on Cityscapes-train and evaluated on Cityscapes-val with image scale [2048, 800] and [2048, 1024]. Default settings are adopted as in MMDetection.}
\label{table:CityScapes}
\end{table}

\begin{table}[h]
\centering
\resizebox{1\linewidth}{!}{
\begin{tabular}{l l l l l l l l l l l l}
\hline
{IoU-Cls} & {IoU-Loc} & {$\text{AP}$} & {$\text{AP}_{50}$} & {$\text{AP}_{75}$} & {$\text{AP}_S$} & {$\text{AP}_M$} & {$\text{AP}_L$} & {$\text{AP}_{60}$} & {$\text{AP}_{70}$} & {$\text{AP}_{80}$} & {$\text{AP}_{90}$}\\
\hline
        &         & 35.7 & 56.8 & 38.5 & 18.8 & 39.7 & 47.4 & 52.1 & 43.9 & 30.6 & 9.3 \\
$\surd$ &         & 36.1 & 57.4 & 38.8 & 19.5 & 40.0 & 48.0 & 52.5 & 44.8 & 31.1 & 9.4 \\
        & $\surd$ & 36.3 & 56.6 & 39.3 & 18.7 & 40.5 & 48.4 & 52.3 & 45.0 & 32.0 & 10.5 \\
$\surd$ & $\surd$ & 36.3 & 57.0 & 39.1 & 19.0 & 40.1 & 48.8 & 52.5 & 44.8 & 31.6 & 10.3\\
\hline
\end{tabular}
}
\caption{ The effectiveness of IoU-balanced classification loss and IoU-balanced localization loss for Faster-RCNN-ResNet50 on COCO \textit{val-2017}.}
\label{table:ablation_fasterrcnn}
\end{table}

\subsection{Discussion}

\textbf{Experimental Results on Two-stage Detector}. IoU-balanced loss functions are general methods and can also be applied to two-stage detector Faster R-CNN. As shown in Table \ref{table:ablation_fasterrcnn}, IoU-balanced classification loss and IoU-balanced localization loss can improve AP by $0.4\%$ and $0.6\%$ respectively. In addition, IoU-balanced localization loss can improve $\text{AP}_{70}\sim\text{AP}_{90}$ by $1.1\%\sim1.4\%$ demonstrating the effectiveness of IoU-balanced localization loss on improving model's localization accuracy. However, combining them gets no further improvement and the improvement for the performance of Faster R-CNN is slightly inferior compared with that of single-stage detectors which demonstrates the problem analyzed in Section \ref{S:1} is slightly alleviated in the two-stage object detector. This is caused by that the regressed proposals generated by RPN in Faster R-CNN have higher localization accuracy compared with the human-designed anchors used in the single-stage detectors. The more accurate proposals used for training Faster R-CNN alleviate the mismatch problem between the classification score and localization accuracy and decrease the number of outliers during training the localization branch.

\section{Conclusions}
\label{S:5}
In this work, we demonstrate that the standard classification loss and localization loss adopted by most of the current single-stage object detectors can severely hurt the model's localization accuracy and thus we propose IoU-balanced loss functions that consist of IoU-balanced classification loss and IoU-balanced localization loss to improve localization accuracy of models. IoU-balanced classification loss is designed to enhance the correlation between classification and localization tasks. And IoU-balanced localization loss is designed to decrease the gradient norm of outliers while increasing the gradient norm of inliers. Extensive experiments on MS COCO, PASCAL VOC and Cityscapes have shown that IoU-balanced loss functions have a substantial improvement for the performance of single-stage detectors, especially for the localization accuracy.

\section{Acknowledgements}
\label{S:6}
This research did not receive any specific grant from funding agencies in the public, commercial, or not-for-profit sectors. Thanks to Dr. Min Lei, Dr. Kaiyou Song and Dr. Xuzhan Chen for their advice and language help.






\bibliographystyle{elsarticle-num-names}
\bibliography{sample.bib}

\begin{thebibliography}{32}
\expandafter\ifx\csname natexlab\endcsname\relax\def\natexlab#1{#1}\fi
\providecommand{\url}[1]{\texttt{#1}}
\providecommand{\href}[2]{#2}
\providecommand{\path}[1]{#1}
\providecommand{\DOIprefix}{doi:}
\providecommand{\ArXivprefix}{arXiv:}
\providecommand{\URLprefix}{URL: }
\providecommand{\Pubmedprefix}{pmid:}
\providecommand{\doi}[1]{\href{http://dx.doi.org/#1}{\path{#1}}}
\providecommand{\Pubmed}[1]{\href{pmid:#1}{\path{#1}}}
\providecommand{\bibinfo}[2]{#2}
\ifx\xfnm\relax \def\xfnm[#1]{\unskip,\space#1}\fi
\bibitem[{Liu et~al.(2016)Liu, Anguelov, Erhan, Szegedy, Reed, Fu, and
  Berg}]{liu2016ssd}
\bibinfo{author}{W.~Liu}, \bibinfo{author}{D.~Anguelov},
  \bibinfo{author}{D.~Erhan}, \bibinfo{author}{C.~Szegedy},
  \bibinfo{author}{S.~Reed}, \bibinfo{author}{C.-Y. Fu}, \bibinfo{author}{A.~C.
  Berg},
\newblock \bibinfo{title}{Ssd: Single shot multibox detector},
\newblock in: \bibinfo{booktitle}{European conference on computer vision},
  \bibinfo{organization}{Springer}, \bibinfo{year}{2016}, pp.
  \bibinfo{pages}{21--37}.
\bibitem[{Redmon et~al.(2016)Redmon, Divvala, Girshick, and
  Farhadi}]{redmon2016you}
\bibinfo{author}{J.~Redmon}, \bibinfo{author}{S.~Divvala},
  \bibinfo{author}{R.~Girshick}, \bibinfo{author}{A.~Farhadi},
\newblock \bibinfo{title}{You only look once: Unified, real-time object
  detection},
\newblock in: \bibinfo{booktitle}{Proceedings of the IEEE conference on
  computer vision and pattern recognition}, \bibinfo{year}{2016}, pp.
  \bibinfo{pages}{779--788}.
\bibitem[{Lin et~al.(2017)Lin, Goyal, Girshick, He, and
  Doll{\'a}r}]{lin2017focal}
\bibinfo{author}{T.-Y. Lin}, \bibinfo{author}{P.~Goyal},
  \bibinfo{author}{R.~Girshick}, \bibinfo{author}{K.~He},
  \bibinfo{author}{P.~Doll{\'a}r},
\newblock \bibinfo{title}{Focal loss for dense object detection},
\newblock in: \bibinfo{booktitle}{Proceedings of the IEEE international
  conference on computer vision}, \bibinfo{year}{2017}, pp.
  \bibinfo{pages}{2980--2988}.
\bibitem[{Zhang et~al.(2018{\natexlab{a}})Zhang, Wen, Bian, Lei, and
  Li}]{zhang2018single_RefineDet}
\bibinfo{author}{S.~Zhang}, \bibinfo{author}{L.~Wen},
  \bibinfo{author}{X.~Bian}, \bibinfo{author}{Z.~Lei}, \bibinfo{author}{S.~Z.
  Li},
\newblock \bibinfo{title}{Single-shot refinement neural network for object
  detection},
\newblock in: \bibinfo{booktitle}{Proceedings of the IEEE Conference on
  Computer Vision and Pattern Recognition}, \bibinfo{year}{2018}{\natexlab{a}},
  pp. \bibinfo{pages}{4203--4212}.
\bibitem[{Zhang et~al.(2018{\natexlab{b}})Zhang, Qiao, Xie, Shen, Wang, and
  Yuille}]{zhang2018single_Entiched}
\bibinfo{author}{Z.~Zhang}, \bibinfo{author}{S.~Qiao},
  \bibinfo{author}{C.~Xie}, \bibinfo{author}{W.~Shen},
  \bibinfo{author}{B.~Wang}, \bibinfo{author}{A.~L. Yuille},
\newblock \bibinfo{title}{Single-shot object detection with enriched
  semantics},
\newblock in: \bibinfo{booktitle}{Proceedings of the IEEE Conference on
  Computer Vision and Pattern Recognition}, \bibinfo{year}{2018}{\natexlab{b}},
  pp. \bibinfo{pages}{5813--5821}.
\bibitem[{Li et~al.(2019)Li, Liu, and Wang}]{li2019gradient}
\bibinfo{author}{B.~Li}, \bibinfo{author}{Y.~Liu}, \bibinfo{author}{X.~Wang},
\newblock \bibinfo{title}{Gradient harmonized single-stage detector},
\newblock in: \bibinfo{booktitle}{Proceedings of the AAAI Conference on
  Artificial Intelligence}, volume~\bibinfo{volume}{33}, \bibinfo{year}{2019},
  pp. \bibinfo{pages}{8577--8584}.
\bibitem[{Ren et~al.(2015)Ren, He, Girshick, and Sun}]{ren2015faster}
\bibinfo{author}{S.~Ren}, \bibinfo{author}{K.~He},
  \bibinfo{author}{R.~Girshick}, \bibinfo{author}{J.~Sun},
\newblock \bibinfo{title}{Faster r-cnn: Towards real-time object detection with
  region proposal networks},
\newblock in: \bibinfo{booktitle}{Advances in neural information processing
  systems}, \bibinfo{year}{2015}, pp. \bibinfo{pages}{91--99}.
\bibitem[{Cai and Vasconcelos(2018)}]{cai2018cascade}
\bibinfo{author}{Z.~Cai}, \bibinfo{author}{N.~Vasconcelos},
\newblock \bibinfo{title}{Cascade r-cnn: Delving into high quality object
  detection},
\newblock in: \bibinfo{booktitle}{Proceedings of the IEEE conference on
  computer vision and pattern recognition}, \bibinfo{year}{2018}, pp.
  \bibinfo{pages}{6154--6162}.
\bibitem[{He et~al.(2017)He, Gkioxari, Doll{\'a}r, and Girshick}]{he2017mask}
\bibinfo{author}{K.~He}, \bibinfo{author}{G.~Gkioxari},
  \bibinfo{author}{P.~Doll{\'a}r}, \bibinfo{author}{R.~Girshick},
\newblock \bibinfo{title}{Mask r-cnn},
\newblock in: \bibinfo{booktitle}{Proceedings of the IEEE international
  conference on computer vision}, \bibinfo{year}{2017}, pp.
  \bibinfo{pages}{2961--2969}.
\bibitem[{Lin et~al.(2017)Lin, Doll{\'a}r, Girshick, He, Hariharan, and
  Belongie}]{lin2017FPN}
\bibinfo{author}{T.-Y. Lin}, \bibinfo{author}{P.~Doll{\'a}r},
  \bibinfo{author}{R.~Girshick}, \bibinfo{author}{K.~He},
  \bibinfo{author}{B.~Hariharan}, \bibinfo{author}{S.~Belongie},
\newblock \bibinfo{title}{Feature pyramid networks for object detection},
\newblock in: \bibinfo{booktitle}{Proceedings of the IEEE conference on
  computer vision and pattern recognition}, \bibinfo{year}{2017}, pp.
  \bibinfo{pages}{2117--2125}.
\bibitem[{Dai et~al.(2016)Dai, Li, He, and Sun}]{dai2016R-FCN}
\bibinfo{author}{J.~Dai}, \bibinfo{author}{Y.~Li}, \bibinfo{author}{K.~He},
  \bibinfo{author}{J.~Sun},
\newblock \bibinfo{title}{R-fcn: Object detection via region-based fully
  convolutional networks},
\newblock in: \bibinfo{booktitle}{Advances in neural information processing
  systems}, \bibinfo{year}{2016}, pp. \bibinfo{pages}{379--387}.
\bibitem[{Girshick(2015)}]{girshick2015fast}
\bibinfo{author}{R.~Girshick},
\newblock \bibinfo{title}{Fast r-cnn},
\newblock in: \bibinfo{booktitle}{Proceedings of the IEEE international
  conference on computer vision}, \bibinfo{year}{2015}, pp.
  \bibinfo{pages}{1440--1448}.
\bibitem[{Girshick et~al.(2014)Girshick, Donahue, Darrell, and
  Malik}]{girshick2014r-cnn}
\bibinfo{author}{R.~Girshick}, \bibinfo{author}{J.~Donahue},
  \bibinfo{author}{T.~Darrell}, \bibinfo{author}{J.~Malik},
\newblock \bibinfo{title}{Rich feature hierarchies for accurate object
  detection and semantic segmentation},
\newblock in: \bibinfo{booktitle}{Proceedings of the IEEE conference on
  computer vision and pattern recognition}, \bibinfo{year}{2014}, pp.
  \bibinfo{pages}{580--587}.
\bibitem[{Bodla et~al.(2017)Bodla, Singh, Chellappa, and
  Davis}]{bodla2017softnms}
\bibinfo{author}{N.~Bodla}, \bibinfo{author}{B.~Singh},
  \bibinfo{author}{R.~Chellappa}, \bibinfo{author}{L.~S. Davis},
\newblock \bibinfo{title}{Soft-nms--improving object detection with one line of
  code},
\newblock in: \bibinfo{booktitle}{Proceedings of the IEEE international
  conference on computer vision}, \bibinfo{year}{2017}, pp.
  \bibinfo{pages}{5561--5569}.
\bibitem[{Gidaris and Komodakis(2015)}]{gidaris2015multiregion}
\bibinfo{author}{S.~Gidaris}, \bibinfo{author}{N.~Komodakis},
\newblock \bibinfo{title}{Object detection via a multi-region and semantic
  segmentation-aware cnn model},
\newblock in: \bibinfo{booktitle}{Proceedings of the IEEE international
  conference on computer vision}, \bibinfo{year}{2015}, pp.
  \bibinfo{pages}{1134--1142}.
\bibitem[{Shrivastava et~al.(2016)Shrivastava, Gupta, and
  Girshick}]{shrivastava2016OHEM}
\bibinfo{author}{A.~Shrivastava}, \bibinfo{author}{A.~Gupta},
  \bibinfo{author}{R.~Girshick},
\newblock \bibinfo{title}{Training region-based object detectors with online
  hard example mining},
\newblock in: \bibinfo{booktitle}{Proceedings of the IEEE conference on
  computer vision and pattern recognition}, \bibinfo{year}{2016}, pp.
  \bibinfo{pages}{761--769}.
\bibitem[{Pang et~al.(2019)Pang, Chen, Shi, Feng, Ouyang, and
  Lin}]{pang2019libraRCNN}
\bibinfo{author}{J.~Pang}, \bibinfo{author}{K.~Chen}, \bibinfo{author}{J.~Shi},
  \bibinfo{author}{H.~Feng}, \bibinfo{author}{W.~Ouyang},
  \bibinfo{author}{D.~Lin},
\newblock \bibinfo{title}{Libra r-cnn: Towards balanced learning for object
  detection},
\newblock in: \bibinfo{booktitle}{Proceedings of the IEEE Conference on
  Computer Vision and Pattern Recognition}, \bibinfo{year}{2019}, pp.
  \bibinfo{pages}{821--830}.
\bibitem[{Tychsen-Smith and Petersson(2018)}]{tychsen2018fitnessNMS}
\bibinfo{author}{L.~Tychsen-Smith}, \bibinfo{author}{L.~Petersson},
\newblock \bibinfo{title}{Improving object localization with fitness nms and
  bounded iou loss},
\newblock in: \bibinfo{booktitle}{Proceedings of the IEEE Conference on
  Computer Vision and Pattern Recognition}, \bibinfo{year}{2018}, pp.
  \bibinfo{pages}{6877--6885}.
\bibitem[{Jiang et~al.(2018)Jiang, Luo, Mao, Xiao, and Jiang}]{jiang2018IoUNet}
\bibinfo{author}{B.~Jiang}, \bibinfo{author}{R.~Luo}, \bibinfo{author}{J.~Mao},
  \bibinfo{author}{T.~Xiao}, \bibinfo{author}{Y.~Jiang},
\newblock \bibinfo{title}{Acquisition of localization confidence for accurate
  object detection},
\newblock in: \bibinfo{booktitle}{Proceedings of the European Conference on
  Computer Vision (ECCV)}, \bibinfo{year}{2018}, pp. \bibinfo{pages}{784--799}.
\bibitem[{Wu et~al.(2019)Wu, Li, and Wang}]{2019IoU-aware}
\bibinfo{author}{S.~Wu}, \bibinfo{author}{X.~Li}, \bibinfo{author}{X.~Wang},
\newblock \bibinfo{title}{Iou-aware single-stage object detector for accurate
  localization},
\newblock \bibinfo{journal}{Image and Vision Computing}
  (\bibinfo{year}{2019}).
\bibitem[{Huang et~al.(2019)Huang, Huang, Gong, Huang, and
  Wang}]{huang2019MSR-CNN}
\bibinfo{author}{Z.~Huang}, \bibinfo{author}{L.~Huang},
  \bibinfo{author}{Y.~Gong}, \bibinfo{author}{C.~Huang},
  \bibinfo{author}{X.~Wang},
\newblock \bibinfo{title}{Mask scoring r-cnn},
\newblock in: \bibinfo{booktitle}{Proceedings of the IEEE Conference on
  Computer Vision and Pattern Recognition}, \bibinfo{year}{2019}, pp.
  \bibinfo{pages}{6409--6418}.
\bibitem[{He et~al.(2015)He, Zhang, Ren, and Sun}]{he2015SPPNet}
\bibinfo{author}{K.~He}, \bibinfo{author}{X.~Zhang}, \bibinfo{author}{S.~Ren},
  \bibinfo{author}{J.~Sun},
\newblock \bibinfo{title}{Spatial pyramid pooling in deep convolutional
  networks for visual recognition},
\newblock \bibinfo{journal}{IEEE transactions on pattern analysis and machine
  intelligence} \bibinfo{volume}{37} (\bibinfo{year}{2015})
  \bibinfo{pages}{1904--1916}.
\bibitem[{Kong et~al.(2020)Kong, Sun, Liu, Jiang, Li, and
  Shi}]{kong2020foveabox}
\bibinfo{author}{T.~Kong}, \bibinfo{author}{F.~Sun}, \bibinfo{author}{H.~Liu},
  \bibinfo{author}{Y.~Jiang}, \bibinfo{author}{L.~Li},
  \bibinfo{author}{J.~Shi},
\newblock \bibinfo{title}{Foveabox: Beyound anchor-based object detection},
\newblock \bibinfo{journal}{IEEE Transactions on Image Processing}
  \bibinfo{volume}{29} (\bibinfo{year}{2020}) \bibinfo{pages}{7389--7398}.
\bibitem[{Lin et~al.(2014)Lin, Maire, Belongie, Hays, Perona, Ramanan,
  Doll{\'a}r, and Zitnick}]{lin2014microCOCO}
\bibinfo{author}{T.-Y. Lin}, \bibinfo{author}{M.~Maire},
  \bibinfo{author}{S.~Belongie}, \bibinfo{author}{J.~Hays},
  \bibinfo{author}{P.~Perona}, \bibinfo{author}{D.~Ramanan},
  \bibinfo{author}{P.~Doll{\'a}r}, \bibinfo{author}{C.~L. Zitnick},
\newblock \bibinfo{title}{Microsoft coco: Common objects in context},
\newblock in: \bibinfo{booktitle}{European conference on computer vision},
  \bibinfo{organization}{Springer}, \bibinfo{year}{2014}, pp.
  \bibinfo{pages}{740--755}.
\bibitem[{Everingham et~al.(2010)Everingham, Van~Gool, Williams, Winn, and
  Zisserman}]{everingham2010pascalVOC}
\bibinfo{author}{M.~Everingham}, \bibinfo{author}{L.~Van~Gool},
  \bibinfo{author}{C.~K. Williams}, \bibinfo{author}{J.~Winn},
  \bibinfo{author}{A.~Zisserman},
\newblock \bibinfo{title}{The pascal visual object classes (voc) challenge},
\newblock \bibinfo{journal}{International journal of computer vision}
  \bibinfo{volume}{88} (\bibinfo{year}{2010}) \bibinfo{pages}{303--338}.
\bibitem[{Cordts et~al.(2016)Cordts, Omran, Ramos, Rehfeld, Enzweiler,
  Benenson, Franke, Roth, and Schiele}]{Cordts2016Cityscapes}
\bibinfo{author}{M.~Cordts}, \bibinfo{author}{M.~Omran},
  \bibinfo{author}{S.~Ramos}, \bibinfo{author}{T.~Rehfeld},
  \bibinfo{author}{M.~Enzweiler}, \bibinfo{author}{R.~Benenson},
  \bibinfo{author}{U.~Franke}, \bibinfo{author}{S.~Roth},
  \bibinfo{author}{B.~Schiele},
\newblock \bibinfo{title}{The cityscapes dataset for semantic urban scene
  understanding},
\newblock in: \bibinfo{booktitle}{Proc. of the IEEE Conference on Computer
  Vision and Pattern Recognition (CVPR)}, \bibinfo{year}{2016}.
\bibitem[{Chen et~al.(2019)Chen, Wang, Pang, Cao, Xiong, Li, Sun, Feng, Liu, Xu
  et~al.}]{chen2019mmdetection}
\bibinfo{author}{K.~Chen}, \bibinfo{author}{J.~Wang},
  \bibinfo{author}{J.~Pang}, \bibinfo{author}{Y.~Cao},
  \bibinfo{author}{Y.~Xiong}, \bibinfo{author}{X.~Li},
  \bibinfo{author}{S.~Sun}, \bibinfo{author}{W.~Feng},
  \bibinfo{author}{Z.~Liu}, \bibinfo{author}{J.~Xu}, et~al.,
\newblock \bibinfo{title}{Mmdetection: Open mmlab detection toolbox and
  benchmark},
\newblock \bibinfo{journal}{arXiv preprint arXiv:1906.07155}
  (\bibinfo{year}{2019}).
\bibitem[{Goyal et~al.(2017)Goyal, Doll{\'a}r, Girshick, Noordhuis, Wesolowski,
  Kyrola, Tulloch, Jia, and He}]{goyal2017accurate}
\bibinfo{author}{P.~Goyal}, \bibinfo{author}{P.~Doll{\'a}r},
  \bibinfo{author}{R.~Girshick}, \bibinfo{author}{P.~Noordhuis},
  \bibinfo{author}{L.~Wesolowski}, \bibinfo{author}{A.~Kyrola},
  \bibinfo{author}{A.~Tulloch}, \bibinfo{author}{Y.~Jia},
  \bibinfo{author}{K.~He},
\newblock \bibinfo{title}{Accurate, large minibatch sgd: Training imagenet in 1
  hour},
\newblock \bibinfo{journal}{arXiv preprint arXiv:1706.02677}
  (\bibinfo{year}{2017}).
\bibitem[{Redmon and Farhadi(2017)}]{redmon2017yolo9000}
\bibinfo{author}{J.~Redmon}, \bibinfo{author}{A.~Farhadi},
\newblock \bibinfo{title}{Yolo9000: better, faster, stronger},
\newblock in: \bibinfo{booktitle}{Proceedings of the IEEE conference on
  computer vision and pattern recognition}, \bibinfo{year}{2017}, pp.
  \bibinfo{pages}{7263--7271}.
\bibitem[{Redmon and Farhadi(2018)}]{redmon2018yolov3}
\bibinfo{author}{J.~Redmon}, \bibinfo{author}{A.~Farhadi},
\newblock \bibinfo{title}{Yolov3: An incremental improvement},
\newblock \bibinfo{journal}{arXiv preprint arXiv:1804.02767}
  (\bibinfo{year}{2018}).
\bibitem[{Dai et~al.(2017)Dai, Qi, Xiong, Li, Zhang, Hu, and
  Wei}]{dai2017deformableConv}
\bibinfo{author}{J.~Dai}, \bibinfo{author}{H.~Qi}, \bibinfo{author}{Y.~Xiong},
  \bibinfo{author}{Y.~Li}, \bibinfo{author}{G.~Zhang}, \bibinfo{author}{H.~Hu},
  \bibinfo{author}{Y.~Wei},
\newblock \bibinfo{title}{Deformable convolutional networks},
\newblock in: \bibinfo{booktitle}{Proceedings of the IEEE international
  conference on computer vision}, \bibinfo{year}{2017}, pp.
  \bibinfo{pages}{764--773}.
\bibitem[{Girshick et~al.(2018)Girshick, Radosavovic, Gkioxari, Doll\'{a}r, and
  He}]{Detectron2018}
\bibinfo{author}{R.~Girshick}, \bibinfo{author}{I.~Radosavovic},
  \bibinfo{author}{G.~Gkioxari}, \bibinfo{author}{P.~Doll\'{a}r},
  \bibinfo{author}{K.~He}, \bibinfo{title}{Detectron},
  \bibinfo{howpublished}{\url{https://github.com/facebookresearch/detectron}},
  \bibinfo{year}{2018}.

\end{thebibliography}







\end{document}